\documentclass[a4paper,final,conference,10pt]{IDAACS}
\usepackage[english]{babel}
\usepackage{amsmath}
\usepackage{amssymb} % Added for mathematical symbols if needed
\usepackage{graphicx}
\usepackage{algorithm}
\usepackage{algpseudocode}
\usepackage{multirow}
\usepackage{cite}
\usepackage{subcaption}
\usepackage{caption}
\usepackage{balance} % To balance columns on the last page
\usepackage{tabularx} % For creating tables with columns that expand to fill a set width
\newcolumntype{Y}{>{\centering\arraybackslash}X} % Defines a new centered, expandable column type
\usepackage{hyperref} % For clickable links/DOIs in references

\hypersetup{
    colorlinks=true,
    linkcolor=blue,
    filecolor=magenta,
    urlcolor=cyan,
    citecolor=green,
    pdfauthor={Svystun et al.},
    pdftitle={YOLO Ensemble for UAV-based Multispectral Defect Detection in Wind Turbine Components}
    }

\title{YOLO Ensemble for UAV-based Multispectral Defect Detection in Wind Turbine Components}

\author{
\IEEEauthorblockN{Serhii Svystun\IEEEauthorrefmark{1},
				    Pavlo Radiuk\IEEEauthorrefmark{1},
                        Oleksandr Melnychenko\IEEEauthorrefmark{1},
                        Oleg Savenko\IEEEauthorrefmark{1},
                        Anatoliy Sachenko\IEEEauthorrefmark{2,3}}

\IEEEauthorblockA{\IEEEauthorrefmark{1}Khmelnytskyi National University, 11, Instytuts’ka str., Khmelnytskyi, Ukraine, 29016,\\
melnychenko@khmnu.edu.ua, radiukp@khmnu.edu.ua, svystuns@khmnu.edu.ua, savenko\_oleg\_st@ukr.net\\
\IEEEauthorrefmark{2}West Ukrainian National University, 11, Lvivska str., Ternopil, Ukraine, 46009, as@wunu.edu.ua\\
\IEEEauthorrefmark{3}Casimir Pulaski Radom University, 29, Malczewskiego str., Radom, Poland, 26-600\\
	}
}

\begin{document}
\maketitle

\begin{abstract}
Unmanned aerial vehicles (UAVs) equipped with advanced sensors have opened up new opportunities for monitoring wind power plants, including blades, towers, and other critical components. However, reliable defect detection requires high-resolution data and efficient methods to process multispectral imagery. In this research, we aim to enhance defect detection accuracy through the development of an ensemble of YOLO-based deep learning models that integrate both visible and thermal channels. We propose an ensemble approach that integrates a general-purpose YOLOv8 model with a specialized thermal model, using a sophisticated bounding box fusion algorithm to combine their predictions. Our experiments show this approach achieves a mean Average Precision (mAP@.5) of 0.93 and an $F_1$-score of 0.90, outperforming a standalone YOLOv8 model, which scored an mAP@.5 of 0.91. These findings demonstrate that combining multiple YOLO architectures with fused multispectral data provides a more reliable solution, improving the detection of both visual and thermal defects.
\end{abstract}


%===========================
\section{Introduction}
Wind power plants play an increasingly critical role in the global transition to renewable energy. However, the efficiency and longevity of wind turbines significantly depend on timely maintenance and the prevention of its components defects~\cite{Sun2022InSitu, Svystun2025DyTAM}. Even minor cracks, corrosion, or temperature anomalies can pose substantial economic and safety risks if left undetected~\cite{Svystun2025Thermal}.

In recent years, unmanned aerial vehicles (UAVs) have emerged as indispensable tools for inspecting large-scale and often remote wind installations. Modern UAVs, equipped with high-resolution optical sensors, thermal cameras, and other specialized imaging devices~\cite{Rizk2024Advanced}, can efficiently collect visual information from multiple angles~\cite{Majumdar2024Enhancing}. The data-gathering capabilities are especially crucial when capturing subtle or internal defects that may not be apparent in standard RGB images alone~\cite{McEnroe2022Survey, Dutta2023Autonomous}.

Although UAV-based inspection holds significant promise, it also introduces complex challenges. Large amounts of data must be processed, often in real-time or near real-time~\cite{Majumdar2024Enhancing}. Environmental conditions such as strong winds, glare, shadows, and varying distances from wind turbine components (WTCs) demand advanced image stabilization and high computational power to identify defects reliably~\cite{Matlekovic2022Microservices}. Given these conditions, robust, intelligent algorithms are essential to sift through multispectral data and accurately detect a wide range of defect types.

The goal of this study is to improve the accuracy and reliability of automated WTC defect detection by developing and evaluating a novel ensemble deep learning approach that leverages fused multispectral (RGB and IR) data acquired via UAVs.

The major contributions of this manuscript are:
\begin{enumerate}
    \item A multispectral image fusion technique that fuses RGB and thermal IR data from UAV inspections, creating enriched image representations that enhance the visibility of diverse defect types.
    \item A novel ensemble learning approach to combining a state-of-the-art YOLOv8 model~\cite{Jocher2023YOLOv8} with a specialized thermal-focused model, utilizing a sophisticated bounding box fusion algorithm to improve overall detection accuracy and robustness.
    \item A comprehensive experimental evaluation demonstrating the proposed ensemble's superior performance against baseline single models and state-of-the-art object detectors.
\end{enumerate}

This paper reviews related work in Section~\ref{sec:relatedWorks} and details our approach in Section~\ref{sec:methods}. We then present our experimental results in Section~\ref{sec:results}, followed by a discussion of the findings in Section~\ref{sec:discussion}. Finally, Section~\ref{sec:conclusion} concludes the paper and outlines future research directions.

%===========================
\section{Related Works}
\label{sec:relatedWorks}
Studies on UAV-based wind turbine monitoring have expanded rapidly~\cite{Sun2022InSitu, Svystun2025Thermal, Matlekovic2022Microservices}, moving beyond early methods that used conventional RGB sensors and basic machine learning~\cite{Deng2021Defect}, which were often limited by complex geometries and environmental conditions. Modern UAVs integrate multiple sensors, including thermal (IR) cameras to detect temperature anomalies indicative of internal flaws~\cite{Svystun2025Thermal, Memari2024DataFusion, McEnroe2022Survey}, and multispectral sensors for identifying issues like early-stage corrosion~\cite{Dutta2023Autonomous, Melnychenko2024Intel}. The resulting high-resolution data streams require efficient processing, leading to the adoption of 5G networks~\cite{McEnroe2022Survey, Dutta2023Autonomous} and edge computing architectures to manage computational overhead~\cite{Majumdar2024Enhancing, Dutta2023Autonomous, Farajzadeh2023SelfEvolving}.

Effective analysis hinges on fusing these different data types~\cite{Melnychenko2024Intel, Farajzadeh2023SelfEvolving, Yang2023Towards}, with techniques evolving from simple averaging~\cite{Zhang2021ImageFusion} to advanced CNNs that learn optimal fusion strategies~\cite{Zhang2020IFCNN, Quan2021FusionNet}. Similarly, detection algorithms have progressed from classifiers with hand-crafted features~\cite{Deng2021Defect, CarmonaTroyo2025Classification, Sun2021Fault} to deep learning models that automatically learn hierarchical defect patterns~\cite{He2016Deep}. Among these, real-time detectors like YOLO offer speed~\cite{Jocher2023YOLOv8, Nguyen2022Performance}, while two-stage detectors such as Faster and Cascade R-CNN provide higher accuracy at a greater computational cost~\cite{Ren2017FasterRCNN, Cai2018CascadeRCNN, Mao2021Automatic, Diaz2023Fast}.

To achieve a balance of speed and accuracy, ensemble techniques have become prominent~\cite{Dietterich2000Ensemble, Zhou2025Ensemble}. These methods can combine fast detectors like YOLO with more precise models like Cascade R-CNN~\cite{Cai2018CascadeRCNN} or EfficientDet~\cite{Tan2020EfficientDet}. Studies confirm that fusing RGB and IR data within such ensembles effectively reduces missed defects, especially in challenging conditions~\cite{Zhou2023Wind, Opitz1997Empirical}, motivating the approach presented in this work.

Overall, research suggests that merging multispectral data with advanced CNN architectures is key to reliable defect detection. However, existing studies often focus on a single detection model or a single spectral range, omitting the potential benefits of ensemble learning.

In this regard, the main objective of this research is to design, implement, and validate an ensemble system based on YOLO architectures, specifically trained to identify cracks, corrosion, and overheating defects on WTCs using a combination of visible and thermal imagery, thereby enhancing the overall performance compared to single-model and single-modality approaches.

%===========================
\section{Methods}
\label{sec:methods}
\begin{figure*}[t]
\centering
\includegraphics[width=\textwidth]{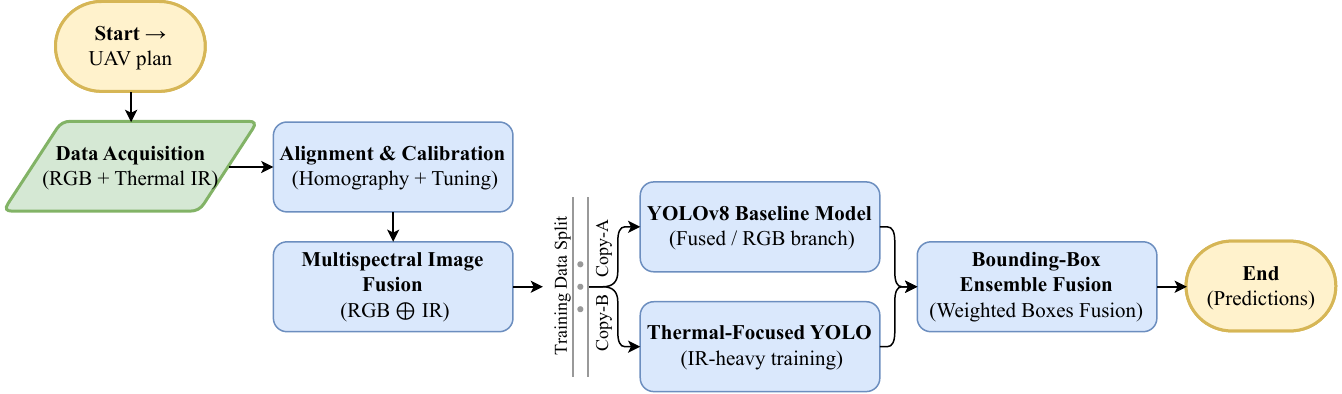}
\caption{General workflow of our proposed ensemble approach. The process involves aligning and fusing RGB and IR data, after which the dataset is used to train a baseline YOLOv8 (Copy-A) and a specialized thermal-focused model $\mathrm{M_t}$ (Copy-B) in parallel. Finally, the predictions from both models are combined using the weighted bounding-box fusion (see Algorithm \ref{alg:ensemble_fusion}) to produce the final, more precise defect detections.}
\label{fig:workflow}
\end{figure*}

This section provides a detailed outline of the proposed approach for constructing and evaluating an ensemble of YOLO-based models for wind turbine defect detection. Our approach is designed to leverage the complementary strengths of multispectral imagery by integrating high-resolution visible (RGB) and thermal (IR) data channels. The overall workflow, as illustrated in Fig.~\ref{fig:workflow}, encompasses four key stages: data acquisition, multispectral image fusion, parallel model training, and a final ensemble fusion stage that combines the outputs into a single, robust prediction.

\subsection{Data Collection and Multispectral fusion}
The foundation of our approach is a rich dataset collected from two primary imaging channels, each providing unique and complementary information about the state of WTCs.
\begin{itemize}
    \item RGB Channel: High-resolution optical images, captured at 4K or 6K, provide detailed textural and color information essential for identifying visual defects such as surface cracks and corrosion on turbine blades, towers, and motor housings.
    \item IR Channel: Thermal images are captured using a co-located infrared camera. This channel is critical for detecting temperature-related anomalies, such as hotspots indicating friction, electrical faults, or subsurface delamination, which are often invisible in the RGB spectrum~\cite{Memari2024DataFusion, Farajzadeh2023SelfEvolving}.
\end{itemize}

A crucial prerequisite for effective fusion is the precise pixel-wise alignment of the IR channel to its corresponding RGB image. To achieve this, we employ a homography-based registration technique, which is well-suited to account for the differences in focal lengths and minor spatial offsets resulting from the physical mounting of the two cameras~\cite{Yang2023Towards, Urtasun2021Pulsed}. The formation of a fused image is formally expressed as:
\begin{equation}
\label{eq:fusion}
\begin{split}
\mathrm{Fusion}(x,y) = & f(\mathrm{RGB}(g_x(x,y), g_y(x,y)), \\
& \mathrm{IR}(h_x(x,y), h_y(x,y))),
\end{split}
\end{equation}
where $g_x,g_y$ and $h_x,h_y$ represent the geometric transformations applied to the RGB and IR images, respectively, and $f(\cdot,\cdot)$ is the fusion function.

Function, formalized in Eq.~(\ref{eq:fusion}), can range from a simple weighted sum for rapid processing to more sophisticated deep learning-based models, such as IFCNN, which can learn an optimal fusion strategy automatically, as illustrated in~\cite{Zhang2020IFCNN}.

\subsection{YOLOv8 Model Training and Additional Specialized Model}
For the core detection task, we selected the YOLOv8 architecture due to its exceptional balance of high-speed inference and state-of-the-art accuracy, making it an ideal candidate for near real-time UAV inspection applications~\cite{Jocher2023YOLOv8}. The models were trained on a comprehensive dataset comprising both standard RGB images and the fused multispectral images to recognize three critical defect classes:
\begin{itemize}
    \item Cracks (C$^1$): Structural fractures in the WTC material, which can compromise aerodynamic integrity and lead to catastrophic failure.
    \item Corrosion (C$^2$): Oxidative degradation on metallic surfaces, primarily on the tower and blade root connections, which weakens structural WTCs over time.
    \item Overheating (C$^3$): Temperature anomalies detected via the IR channel, often indicative of failing mechanical or electrical WTCs within the nacelle.
\end{itemize}

\subsection{Ensemble Fusion Algorithm}
\label{subsec:ensemble_fusion}
The final step of our approach is to intelligently combine the predictions from the baseline YOLOv8 model ($O_y$) and the specialized thermal model ($O_{\mathrm{M_t}}$). For this, we developed a bounding box fusion algorithm designed to merge overlapping detections of the same class while preserving unique findings from each model. The output of the baseline model is defined as $O_y = \{ D_{y,i} = (c_{y,i}, b_{y,i}, p_{y,i}) \}_{i=1}^{N_y}$ and the output of the specialized model is $O_{\mathrm{M_t}} = \{ D_{\mathrm{M_t},j} = (c_{\mathrm{M_t},j}, b_{\mathrm{M_t},j}, p_{\mathrm{M_t},j}) \}_{j=1}^{N_{\mathrm{M_t}}}$, where $c$ represents the predicted class, $b = (x, y, W, H)$ is the bounding box, and $p$ is the confidence score.

When the models identify overlapping bounding boxes of the same class, a new fused detection, $D_{\text{ensemble}}$, is created by calculating weighted averages of their confidence scores and bounding box parameters:
\begin{align}
    p_{\text{ensemble}} &= \gamma p_{\mathrm{M_t},j} + (1-\gamma) p_{y,i}, \label{eq:conf_fuse} \\
    x_{\text{ensemble}} &= \gamma x_{\mathrm{M_t},j} + (1-\gamma) x_{y,i}, \label{eq:x_fuse} \\
    y_{\text{ensemble}} &= \gamma y_{\mathrm{M_t},j} + (1-\gamma) y_{y,i}, \label{eq:y_fuse} \\
    W_{\text{ensemble}} &= \gamma W_{\mathrm{M_t},j} + (1-\gamma) W_{y,i}, \label{eq:w_fuse} \\
    H_{\text{ensemble}} &= \gamma H_{\mathrm{M_t},j} + (1-\gamma) H_{y,i}. \label{eq:h_fuse}
\end{align}

The weighting factor, $\gamma \in [0, 1]$ from Eqs.~(\ref{eq:conf_fuse})--(\ref{eq:h_fuse}), controls the contribution of each model, where a value of $\gamma=0.5$ assigns equal importance. After this fusion step, the combined list of all fused and unmerged detections undergoes a final, critical post-processing stage: Non-Maximum Suppression (NMS). NMS iteratively selects the box with the highest confidence score and suppresses any other boxes that have an Intersection over Union (IoU) with it above a predefined threshold ($\tau_{\text{nms}}$). The complete procedure is formally detailed in Algorithm~\ref{alg:ensemble_fusion}.

Algorithm~\ref{alg:ensemble_fusion} leverages the consensus between the two models for high-confidence detections while preserving unique findings from each, thereby aiming for improved precision. To ensure the reproducibility of our findings and to encourage further research, the complete source code for the cyber-physical system is publicly available in a GitHub repository~\cite{Svystun2025UAVRepo}.

\begin{algorithm}[!tp]
\caption{Bounding Box Ensemble Fusion}
\label{alg:ensemble_fusion}
\begin{algorithmic}[1]
\Require
$O_y, O_{\mathrm{M_t}}$: Detections from baseline and thermal models.
$\tau_{\text{iou}}, \gamma, \tau_{\text{nms}}$: IoU, fusion weight, and NMS thresholds.
\Ensure $O_{\text{final}}$: Final set of fused detections.

\State $O_{\text{fused}}, O_{\text{unmerged}} \gets \emptyset, \emptyset$
\State Initialize \texttt{merged\_flags} for all detections in $O_y, O_{\mathrm{M_t}}$ to \texttt{false}.

\For{each detection $D_i \in O_y$}
    \State \texttt{match\_found} $\gets$ \texttt{false}
    \For{each detection $D_j \in O_{\mathrm{M_t}}$}
        \If{not merged($D_i, D_j$) and $D_i.\text{class} = D_j.\text{class}$ and $\text{IoU}(D_i.\text{box}, D_j.\text{box}) \ge \tau_{\text{iou}}$}
            \State $D_{\text{fused}} \gets \text{Fuse}(D_i, D_j, \gamma)$ by Eqs.~(\ref{eq:conf_fuse})--(\ref{eq:h_fuse}).
            \State $O_{\text{fused}} \gets O_{\text{fused}} \cup \{D_{\text{fused}}\}$.
            \State Mark $D_i, D_j$ as \texttt{merged}; \texttt{match\_found} $\gets$ \texttt{true}; \textbf{break}.
        \EndIf
    \EndFor
    \If{not \texttt{match\_found}} $O_{\text{unmerged}} \gets O_{\text{unmerged}} \cup \{D_i\}$. \EndIf
\EndFor

\State Add all remaining unmerged detections from $O_{\mathrm{M_t}}$ to $O_{\text{unmerged}}$.
\State $O_{\text{combined}} \gets O_{\text{fused}} \cup O_{\text{unmerged}}$.
\State $O_{\text{final}} \gets \text{NMS}(O_{\text{combined}}, \tau_{\text{nms}})$.
\State \Return $O_{\text{final}}$
\end{algorithmic}
\end{algorithm}

%===========================
\section{Results}
\label{sec:results}
This section presents the quantitative and qualitative findings of our proposed approach. The performance of the single YOLOv8 model and the proposed ensemble was assessed on a comprehensive test set.

\subsection{Experimental Setup}
Our training dataset was built upon the publicly available Blade30 dataset~\cite{Yang2023Towards}, which contains 1,102 images featuring cracks and corrosion. To enhance the model's capabilities, we augmented this dataset with 670 additional images specifically capturing overheating events. The training process employed the Adam optimizer with a learning rate of $\alpha = 0.001$ and an early-stopping criterion to prevent overfitting. To further improve detection robustness, we trained a second, specialized model, denoted $\mathrm{M_t}$. This model was trained exclusively on IR-heavy data to develop a heightened sensitivity to subtle thermal gradients that might be overlooked by a general-purpose model processing fused data~\cite{McEnroe2022Survey, Dutta2023Autonomous}.

Detection accuracy was measured using standard object detection metrics to ensure a robust and fair comparison. These include: (i) mean Average Precision mAP@.5, (ii) mAP@.5:.95, (iii) precision, and (iv) $F_1$-Score.

\subsection{Performance of Single Model vs. Ensemble}
The investigation was to determine the performance uplift gained by moving from a single, highly-capable detector to a specialized ensemble, with a detailed comparison provided in Table~\ref{tab:combined_results}. The single YOLOv8 model established a strong baseline, achieving a mean mAP@.5 of 0.91 and a mean $F_1$-score of 0.88, demonstrating proficiency across the defined defect classes. Our proposed ensemble approach yielded consistent improvements, elevating the mean mAP@.5 to 0.93 and the stricter mAP@.5:.95 to 0.79. This represents a 2.2\% relative improvement in mAP@.5 and a 3.9\% gain in the more demanding mAP@.5:.95 metric.

The gains were particularly pronounced for specific defect classes. For ``Cracks'' (C$^1$), the $F_1$-score improved from 0.90 to 0.92. For the ``Overheating'' (C$^3$) class, the $F_1$-score rose from 0.86 to 0.89, underscoring the value of the specialized thermal model ($\mathrm{M_t}$), whose expertise in identifying thermal anomalies compensates for the limitations of a general-purpose detector. The ensemble's ability to reduce both false negatives and false positives confirms its superior robustness and reliability for this application.

\begin{table*}[!tp]
\caption{Comparative Performance of Single YOLOv8 Model vs. Proposed Ensemble (YOLOv8 + $\mathrm{M_t}$) on the Multispectral Test Set. Bold Text Represents Higher Values.}
\label{tab:combined_results}
\centering
\renewcommand{\arraystretch}{1.3} % Adjusts row height for better readability
% The column specifier |l|*{8}{Y|} creates a left-aligned column and 8 expandable, centered columns, all fully bordered.
\begin{tabularx}{\textwidth}{|l|*{8}{Y|}}
\hline
\multirow{2}{*}{\textbf{Defect Class}} & \multicolumn{2}{c|}{\textbf{mAP@.5}} & \multicolumn{2}{c|}{\textbf{mAP@.5:.95}} & \multicolumn{2}{c|}{\textbf{Precision}} & \multicolumn{2}{c|}{\textbf{$F_1$-Score}} \\
\cline{2-9}
& \textbf{YOLOv8} & \textbf{Ensemble} & \textbf{YOLOv8} & \textbf{Ensemble} & \textbf{YOLOv8} & \textbf{Ensemble} & \textbf{YOLOv8} & \textbf{Ensemble} \\
\hline
Cracks (C$^1$)      & 0.93 & \textbf{0.95} & 0.78 & \textbf{0.81} & 0.91 & \textbf{0.93} & 0.90 & \textbf{0.92} \\
\hline
Corrosion (C$^2$)   & 0.90 & \textbf{0.92} & 0.75 & \textbf{0.78} & 0.88 & \textbf{0.90} & 0.87 & \textbf{0.89} \\
\hline
Overheating (C$^3$) & 0.89 & \textbf{0.92} & 0.74 & \textbf{0.77} & 0.87 & \textbf{0.90} & 0.86 & \textbf{0.89} \\
\hline
Mean & 0.91 & \textbf{0.93} & 0.76 & \textbf{0.79} & 0.89 & \textbf{0.91} & 0.88 & \textbf{0.90} \\
\hline
\end{tabularx}
\end{table*}

\subsection{Comparison with State-of-the-Art Approaches}
Next, we evaluated the proposed ensemble against several well-established object detection architectures: Faster R-CNN~\cite{Ren2017FasterRCNN}, Cascade R-CNN~\cite{Cai2018CascadeRCNN}, and EfficientDet~\cite{Tan2020EfficientDet}. For a direct comparison, all models were trained on the same fused multispectral dataset, with the results summarized in Table~\ref{tab:compSOTA}.

The evaluation demonstrates that while two-stage detectors are robust, with Faster R-CNN and Cascade R-CNN achieving mAP@.5 scores of 0.89 and 0.90 respectively, modern single-stage models are more effective for this task. Both EfficientDet-D1 and our baseline single YOLOv8 model reached an mAP@.5 of 0.91.

Our proposed ensemble surpassed all other models, delivering a top-tier mAP@.5 of 0.93 and an mAP@.5:.95 of 0.79. This performance uplift underscores the value of our approach. The ensemble capitalizes on YOLOv8’s feature extraction power while leveraging the specialized thermal model ($\mathrm{M_t}$) to accurately detect defects with subtle thermal signatures.

\begin{table}[htb]
\caption{Comparative Performance with State-of-the-Art Models on Our Test Set. Bold Text Means Higher Values.}
\label{tab:compSOTA}
\centering
\begin{tabular}{|l|c|c|}
\hline
\textbf{Model} & \textbf{mAP@.5} & \textbf{mAP@.5:.95} \\
\hline
Faster R-CNN~\cite{Ren2017FasterRCNN} & 0.89 & 0.74 \\
\hline
Cascade R-CNN~\cite{Cai2018CascadeRCNN} & 0.90 & 0.75 \\
\hline
EfficientDet-D1~\cite{Tan2020EfficientDet} & 0.91 & 0.76 \\
\hline
YOLOv8 (Single Model)~\cite{Jocher2023YOLOv8} & 0.91 & 0.76 \\
\hline
Our Ensemble (YOLOv8 + $\mathrm{M_t}$) & \textbf{0.93} & \textbf{0.79} \\
\hline
\end{tabular}
\end{table}

\subsection{Qualitative Analysis}
\begin{figure*}[!tp]
    \centering

    % --- First Row of Images ---
    \begin{subfigure}[b]{0.23\textwidth}
        \includegraphics[width=\linewidth]{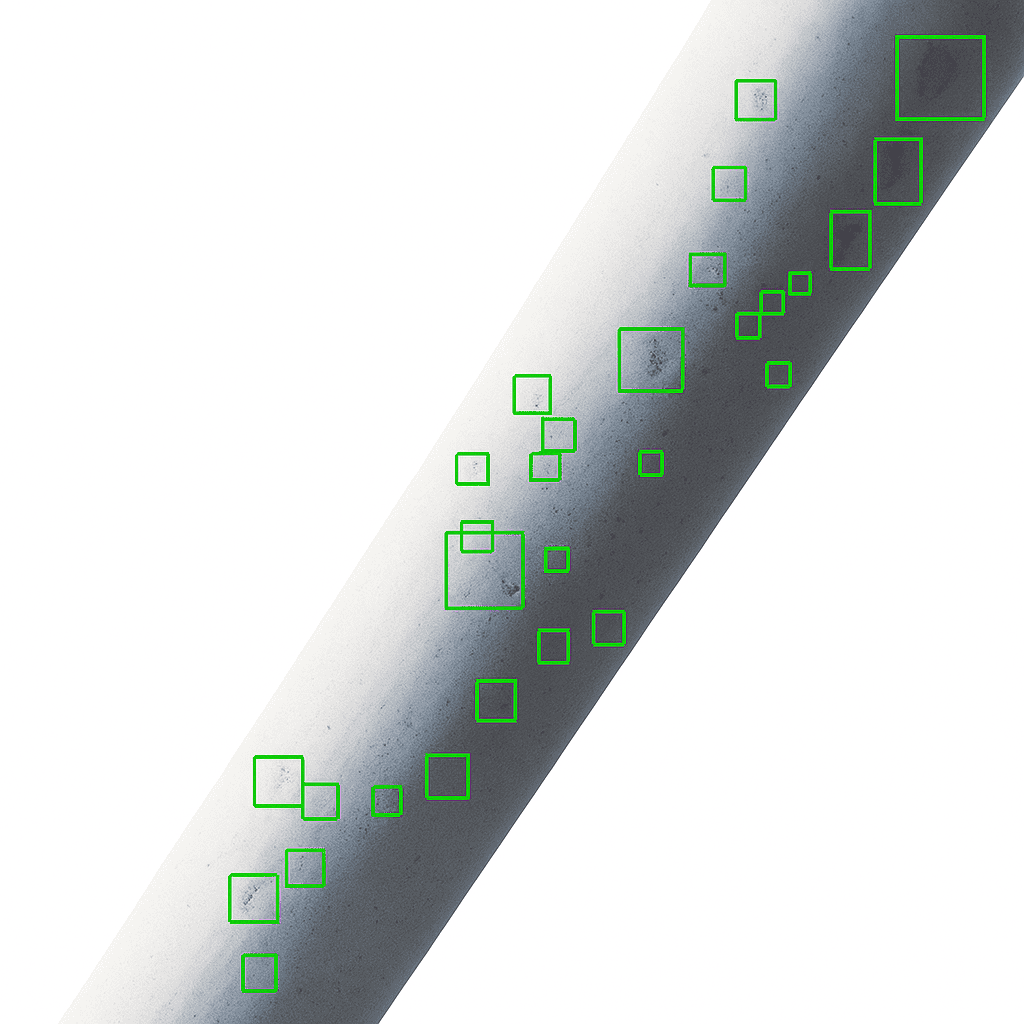}
        \caption{}
        \label{fig:qualitative_results_a}
    \end{subfigure}
    \hfill % Distributes horizontal space
    \begin{subfigure}[b]{0.23\textwidth}
        \includegraphics[width=\linewidth]{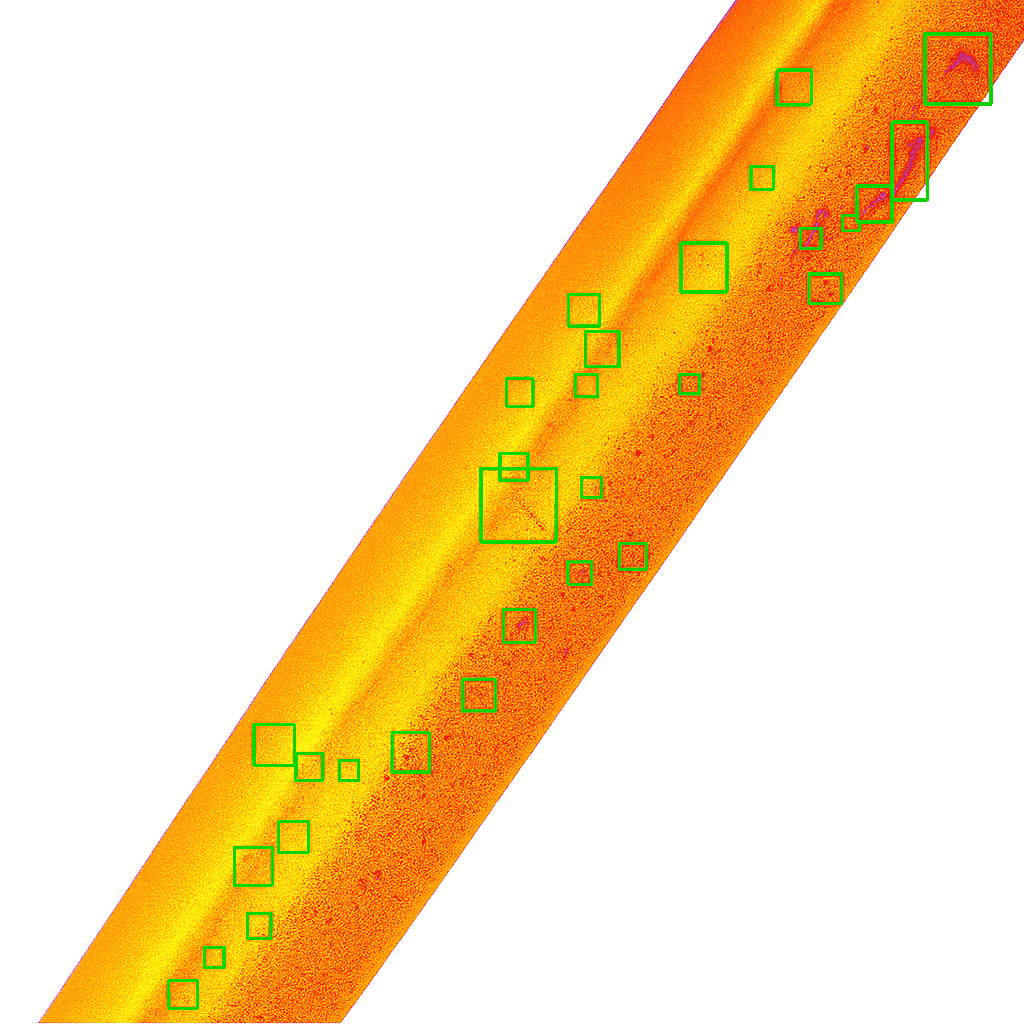}
        \caption{}
        \label{fig:qualitative_results_b}
    \end{subfigure}
    \hfill
    \begin{subfigure}[b]{0.23\textwidth}
        \includegraphics[width=\linewidth]{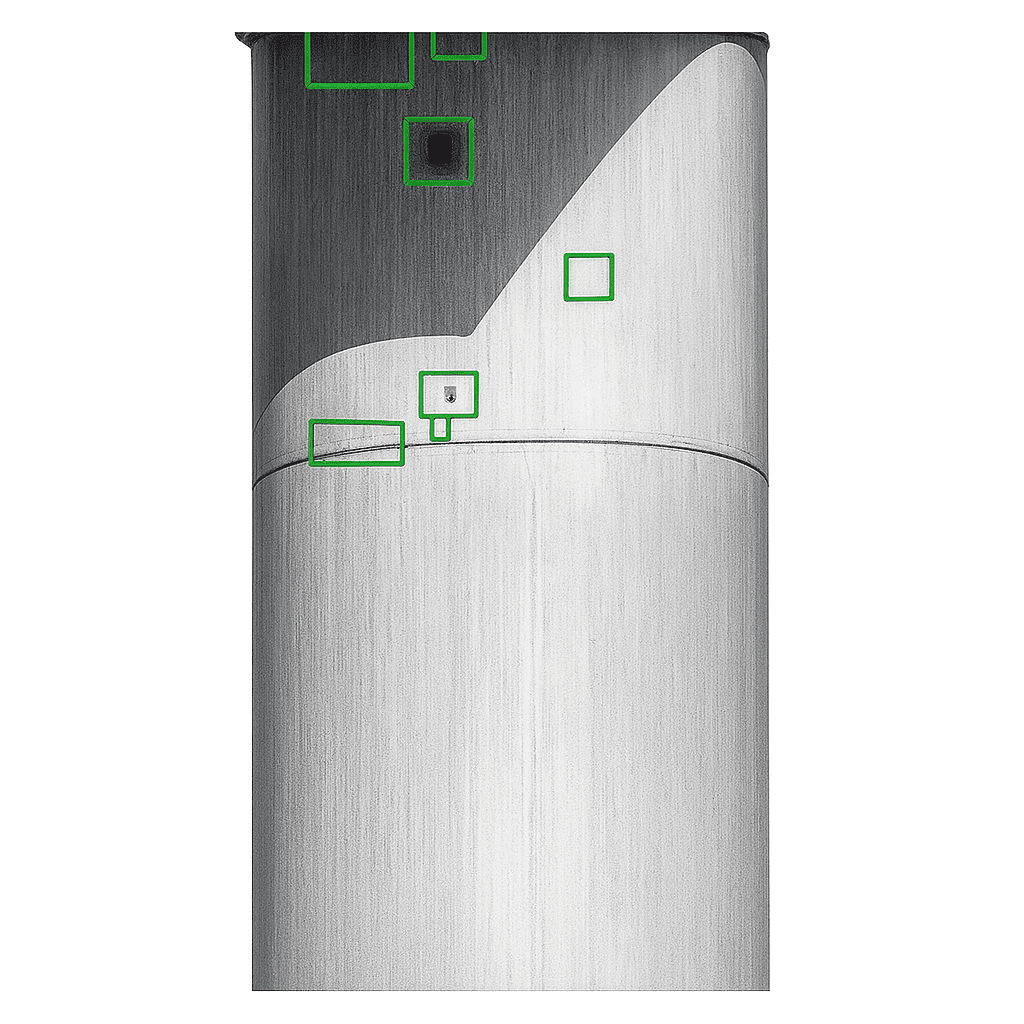}
        \caption{}
        \label{fig:qualitative_results_c}
    \end{subfigure}
    \hfill
    \begin{subfigure}[b]{0.23\textwidth}
        \includegraphics[width=\linewidth]{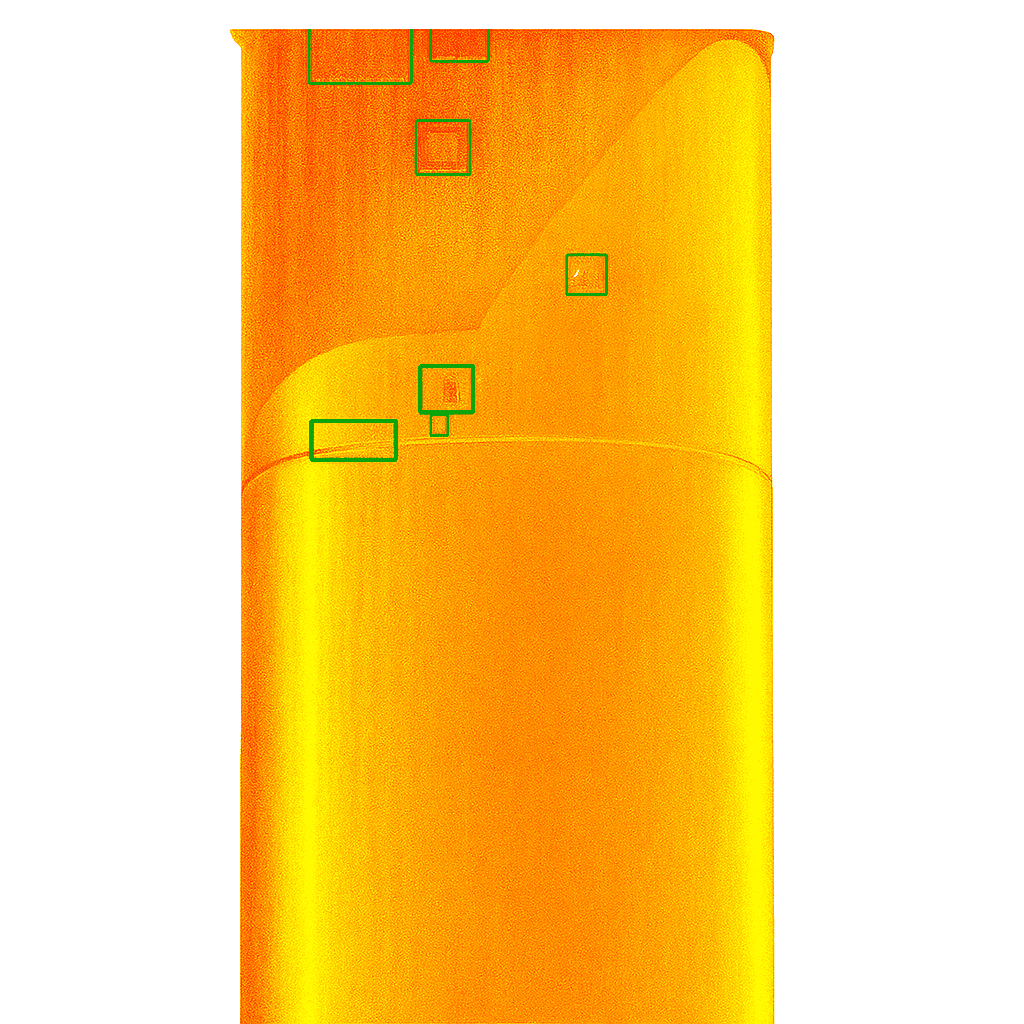}
        \caption{}
        \label{fig:qualitative_results_d}
    \end{subfigure}

    \vspace{\floatsep} % Adds a standard vertical space between rows

    % --- Second Row of Images ---
    \begin{subfigure}[b]{0.23\textwidth}
        \includegraphics[width=\linewidth]{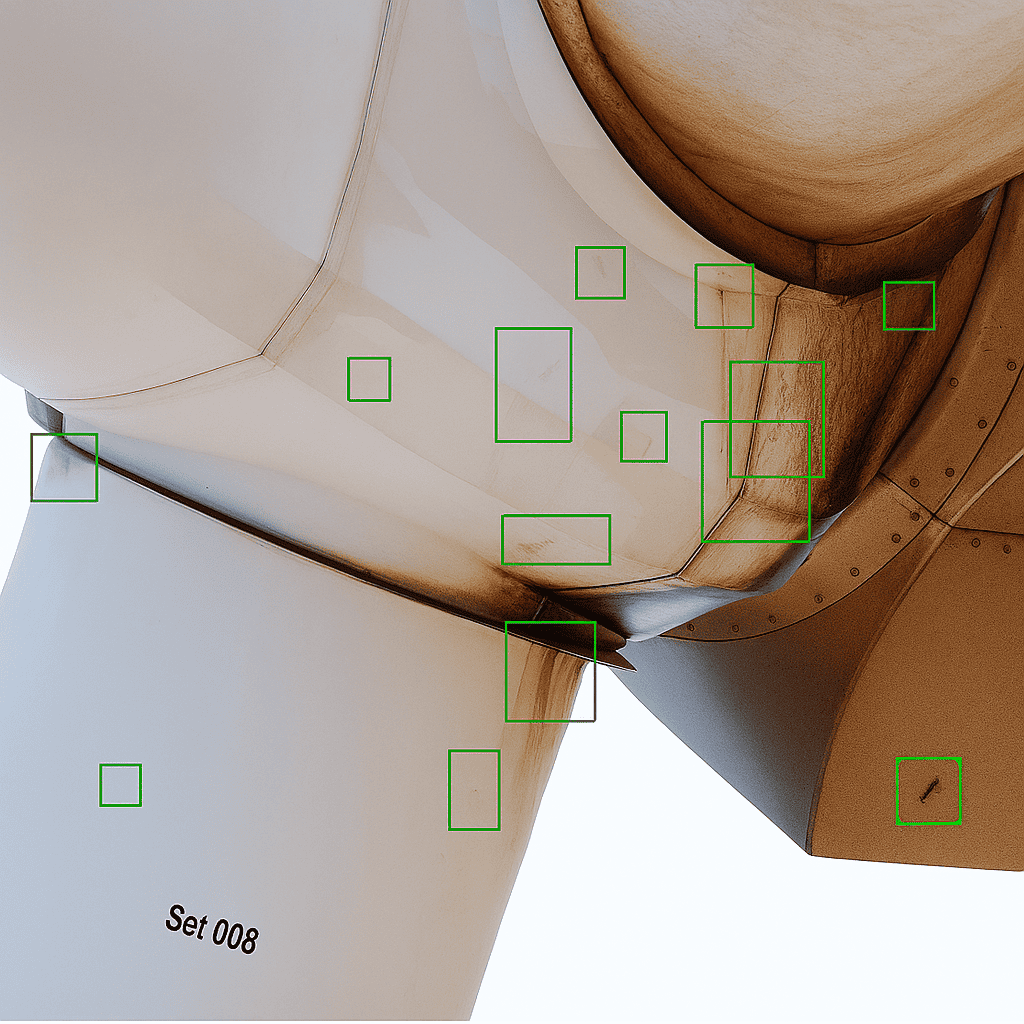}
        \caption{}
        \label{fig:qualitative_results_e}
    \end{subfigure}
    \hfill
    \begin{subfigure}[b]{0.23\textwidth}
        \includegraphics[width=\linewidth]{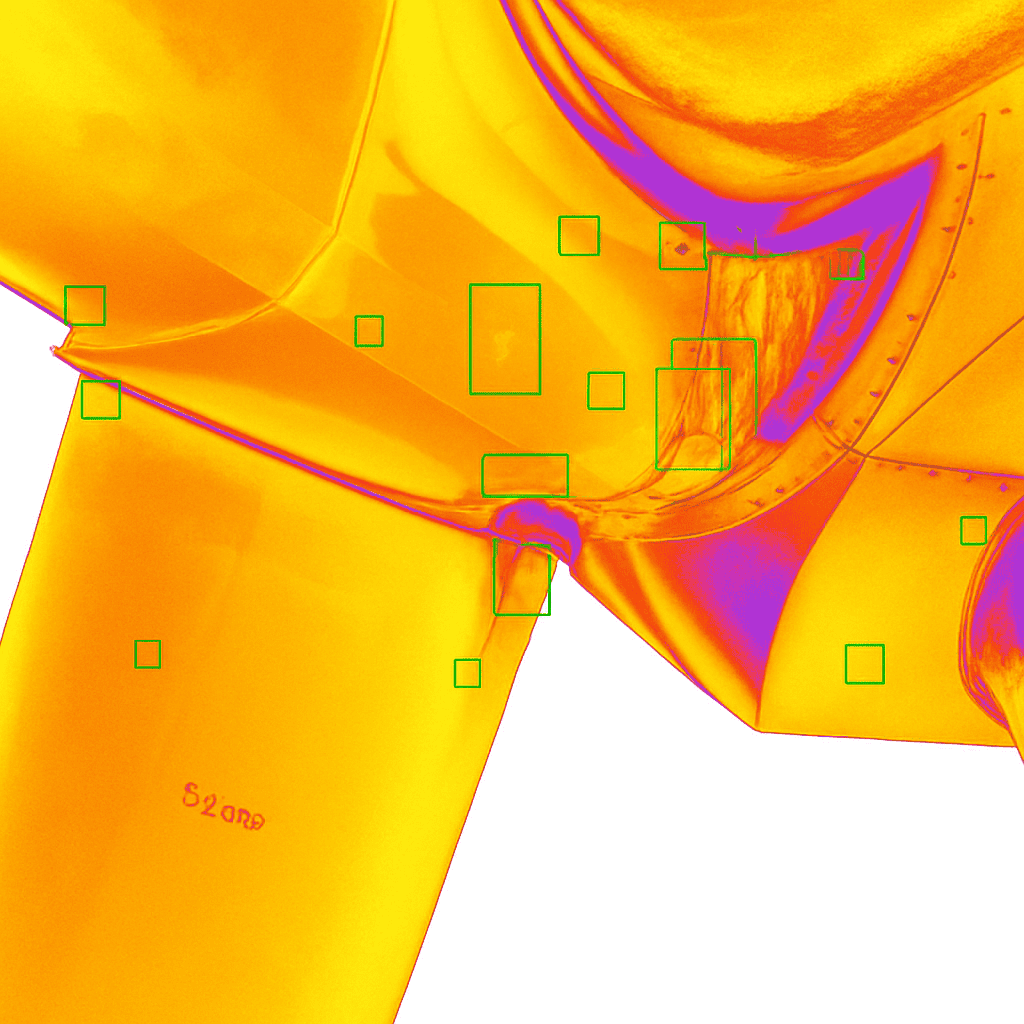}
        \caption{}
        \label{fig:qualitative_results_f}
    \end{subfigure}
    \hfill
    \begin{subfigure}[b]{0.23\textwidth}
        \includegraphics[width=\linewidth]{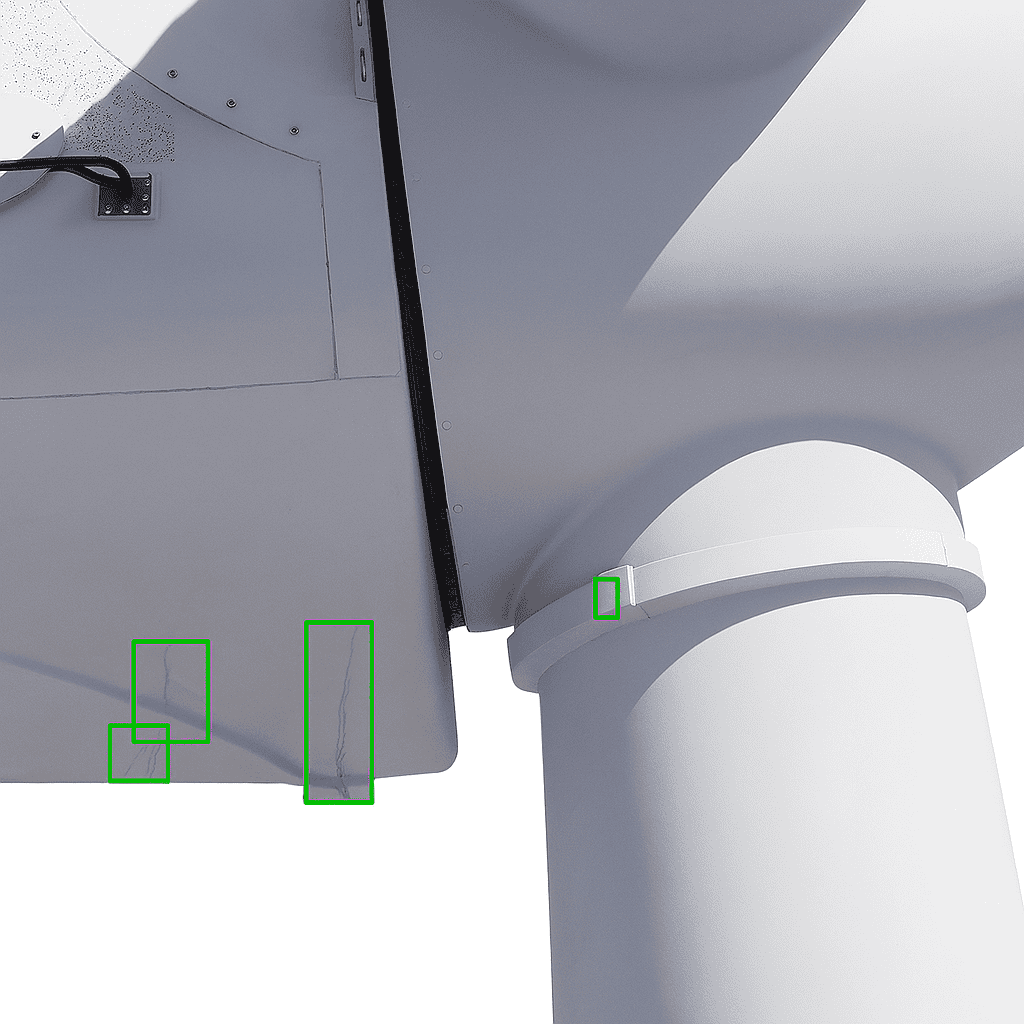}
        \caption{}
        \label{fig:qualitative_results_g}
    \end{subfigure}
    \hfill
    \begin{subfigure}[b]{0.23\textwidth}
        \includegraphics[width=\linewidth]{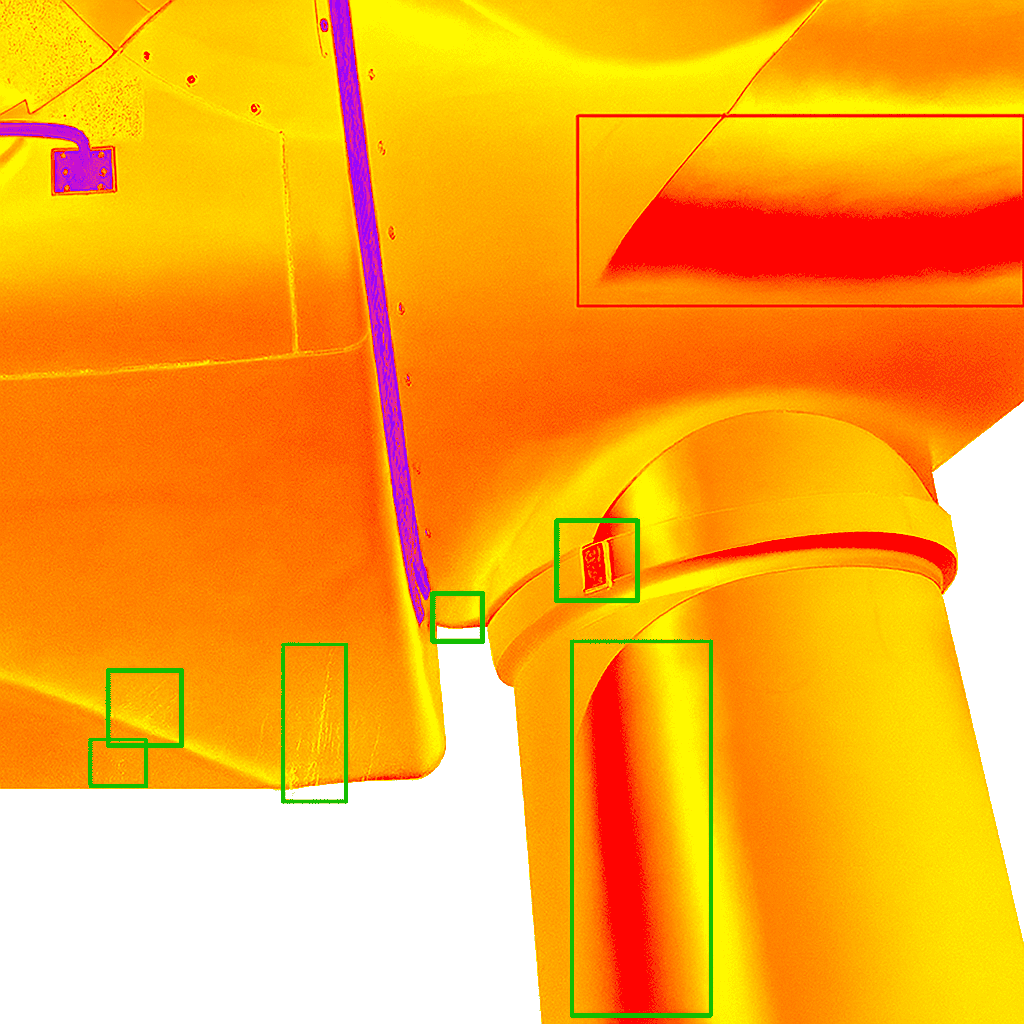}
        \caption{}
        \label{fig:qualitative_results_h}
    \end{subfigure}

    \caption{Qualitative validation of the proposed multispectral fusion across key wind turbine components. Our approach enhances the detection of visible surface defects, such as cracks on the blade (\subref{fig:qualitative_results_a}, \subref{fig:qualitative_results_b}) and corrosion on the tower (\subref{fig:qualitative_results_c}, \subref{fig:qualitative_results_d}), by improving contrast. Furthermore, it excels at identifying non-visible thermal anomalies, revealing potential stress points on the rotor (\subref{fig:qualitative_results_e}, \subref{fig:qualitative_results_f}) and critical overheating in the motor component (\subref{fig:qualitative_results_g}, \subref{fig:qualitative_results_h}) that are completely missed by RGB-only inspection.}
    \label{fig:qualitative_results}
\end{figure*}

To complement the quantitative metrics, Fig.~\ref{fig:qualitative_results} presents a visual comparison of defect detection on standard RGB images versus the fused multispectral images generated by our system. The benefit of fusion is evident for visible defects. For the turbine blade, the standard RGB image (Fig.~\ref{fig:qualitative_results}\subref{fig:qualitative_results_a}) shows a crack, but its visibility is hampered by shadows and surface texture. The fused image (Fig.~\ref{fig:qualitative_results}\subref{fig:qualitative_results_b}) noticeably enhances the contrast along the fracture line, allowing the ensemble model to generate a tighter, higher-confidence bounding box. Similarly, for the tower, the fused image (Fig.~\ref{fig:qualitative_results}\subref{fig:qualitative_results_d}) improves the detection of corrosion compared to the original (Fig.~\ref{fig:qualitative_results}\subref{fig:qualitative_results_c}) by making the subtle surface degradation more distinct against the background.

The system's strength is particularly notable for identifying defects that are not visible in standard imagery. On the rotor hub, while the RGB view (Fig.~\ref{fig:qualitative_results}\subref{fig:qualitative_results_e}) appears unremarkable, the fusion process reveals subtle thermal patterns indicative of structural stress (Fig.~\ref{fig:qualitative_results}\subref{fig:qualitative_results_f}). Most critically, the RGB image of the motor component shows no obvious signs of damage (Fig.~\ref{fig:qualitative_results}\subref{fig:qualitative_results_g}), whereas the fused image clearly reveals distinct thermal hotspots (Fig.~\ref{fig:qualitative_results}\subref{fig:qualitative_results_h})—a critical defect that a standard visual inspection would otherwise miss.

%===========================
\section{Discussion}
\label{sec:discussion}
The experimental results confirm our ensemble learning approach advances automated wind turbine inspection. By integrating a general-purpose YOLOv8 model with a specialized thermal model ($\mathrm{M_t}$), our approach achieves a mean $F_1$-score of 0.90, outperforming single-model architectures. This work builds upon prior studies that validated RGB-IR fusion~\cite{Zhou2023Wind, Memari2024DataFusion} by introducing a structured ensemble that systematically combines the speed of single-stage detectors with the precision of a specialist model, an improvement over using single complex architectures like Cascade R-CNN~\cite{Mao2021Automatic}.

The primary advantage is the enhanced detection of defects often missed in RGB-only imagery, such as subtle cracks obscured by shadows and overheating anomalies invisible in the visible spectrum~\cite{Deng2021Defect}. This multi-modal synergy provides a more comprehensive diagnostic tool, directly addressing the limitations of systems that rely on a single spectral range. However, this improved accuracy introduces greater computational complexity and longer inference times, which presents a trade-off for real-time deployment on computationally constrained UAVs.

Despite promising results, the system’s performance is dependent on precise sensor alignment and can be affected by environmental noise in thermal data. Future research should focus on mitigating these challenges by optimizing the ensemble for edge computing, developing more robust sensor fusion algorithms, and expanding the approach to incorporate additional data modalities, such as hyperspectral imaging as explored in~\cite{Rizk2024Advanced}.

%===========================
\section{Conclusion}
\label{sec:conclusion}
In this study, we developed and validated a novel ensemble-based deep learning approach for high-precision defect detection in wind power plants using fused multispectral UAV imagery. By integrating a baseline YOLOv8 model with a specialized thermal-focused model via a sophisticated bounding box fusion algorithm, our system enhances diagnostic accuracy over single-model approaches. Experimental results underscore its effectiveness, achieving a mean $F_1$-score of 0.90 and a mean Average Precision (mAP@.5) of 0.93. The integration of thermal data proved particularly impactful, enabling the identification of overheating, subsurface cracks, and corrosion often missed in RGB-only inspections. While limited by computational overhead, sensor alignment requirements, and environmental sensitivity, our work establishes a more robust approach to automated inspection.

Future research will focus on optimizing the ensemble for real-time edge deployment, incorporating more advanced data fusion techniques to mitigate noise and misregistration.

%===========================
\section*{Declaration on Generative AI}
\label{sec:declaration}
During the preparation of this work, the authors utilized generative AI tools (Gemini v2.5 Pro and Grammarly) to assist with grammar checks, spelling corrections, and paraphrasing. After using these services, the authors reviewed and edited the content as needed and take full responsibility for the publication's content.

\balance % Balance columns on the last page
\bibliographystyle{IEEEtran}
\bibliography{IEEEabrv,Svystun_2025_Preprint}

\end{document}